\documentclass[11pt]{article}

\usepackage[preprint]{wmt/acl}

\usepackage{times}
\usepackage{latexsym}

\usepackage[T1]{fontenc}
\usepackage[utf8]{inputenc}
\usepackage{microtype}
\usepackage{inconsolata}

\usepackage{custom}

\title{\facet at WMT 2026 Automated Translation Quality Evaluation Task}

\author{
  Ahrii Kim\textsuperscript{1} \quad
  Chanjun Park\textsuperscript{2 $\star$} \quad
  Seong-heum Kim\textsuperscript{1,3 $\star$} \\
  \textsuperscript{1}AI-Bio Convergence Research Inst. \quad
  \textsuperscript{2}School of Software \quad
  \textsuperscript{3}Dept. of Intelligent Semiconductors \quad
  \\ Soongsil University \\
  \texttt{\{ahriikim,chanjun.park,seongheum\}@ssu.ac.kr} \\
}

\begin{document}
\maketitle

\begingroup
\renewcommand{\thefootnote}{$\star$}
\footnotetext{Corresponding authors.}
\endgroup

\blfootnote{\faGithub{} \url{https://github.com/trotacodigos/facet.git}}

\begin{abstract}
Different error types in machine translation require different evidence. Whether meaning is
preserved can be judged only against the source, while whether the target is well-formed, or
whether it names one entity consistently, can be judged from the target alone. We present
\facet, our reference-free submission to the WMT26 Automated Translation Quality Evaluation Task, which decomposes evaluation into \flu, \acc, and \cst passes and gives each
pass only the context its error type requires. A single fixed model is prompted three times,
and the merged error spans yield the three task outputs, error spans, quality scores, and
error-free labels, with no trained components. We also submit \facetc, which omits the
\cst pass. Without gold labels, we characterize the predictions of \facet. Its system
rankings place post-edited human translation first, and the \cst pass changes about a
tenth of segment scores while leaving the ranking nearly unchanged.
\end{abstract}

\section{Introduction}
\label{sec:intro}
Judging the quality of a translation is not one decision but several, and they do not rest
on the same evidence. Whether the target preserves the meaning of the source can be decided
only from the source and the target together. Whether the target is well-formed is a
property of the target as a piece of language, and whether a recurring name or term keeps
one rendering is a property of the target across a whole document. The source contributes
nothing to the last two questions, and a single segment is too short to answer the third.
LLM-based evaluators typically put all of these questions to the model in one prompt over
one input \citep{kocmi-federmann-2023-gemba,fernandes-etal-2023-devil,lu-etal-2024-error,
junczys-dowmunt-2025-gemba}, so every judgment is made from the same context whether or not it is the context
that judgment needs.

\facet (\textbf{F}luency, \textbf{A}ccuracy, and \textbf{C}onsistency \textbf{E}valuation of
\textbf{T}ranslations) separates the judgments and the inputs. Its three passes follow the
MQM error taxonomy \citep{burchardt-2013-multidimensional}. \acc and \flu are the two
top-level MQM dimensions, with terminology, style, and locale conventions folded into \flu
as target-only judgments. \cst is an MQM subtype of \flu that we promote to a pass of its
own, because it is the one \flu judgment that cannot be made within a segment.

The passes run under asymmetric input. The \acc pass reads the source and the target. The
\flu and \cst passes read the target only, each at the scope its error type requires, the
segment for \flu and the whole document for \cst. The \cst pass judges only recurring
items, that is, named entities, domain terms, and fixed expressions. It checks that one
entity does not appear under two or more names a reader would take as different referents,
and that shortened or pronominal references agree with the form they refer back to. Because
it judges internal consistency rather than correctness, it needs no source, and the
restriction to recurring items limits overlap with the other two passes.

\facet is our submission to the WMT26 shared task on Automated Translation Quality, which merges the metrics and quality estimation (QE) tracks of previous years.
The task provides reference translations as an optional input and scores segments in
context, where the context is a document or a long multi-segment unit. \facet uses no
references and treats the document context as an input in its own right. The system has no
trained or fine-tuned components. A single fixed model is prompted, and the behavior of the
system is determined by the decomposition into passes and by the input each pass receives.
One merged prediction yields the three shared-task outputs, error span annotation, quality
scoring, and error-free detection.

Our primary submission, \facet, uses all three passes. Whether a document-level judgment
pays off under segment-level scoring is an open question, so we also submit \facetc, an
ablation that omits \cst, and use the pair to measure what the document-level pass
contributes.
\section{Related Work}
\label{sec:relwork}
\paragraph{LLM-as-judge evaluation.}
Prompting large language models (LLMs) for translation quality assessment is now common
practice. GEMBA \citep{kocmi-federmann-2023-large} prompts GPT models for direct quality
scores, and GEMBA-MQM \citep{kocmi-federmann-2023-gemba} extends this to reference-free
MQM-style error annotation \citep{burchardt-2013-multidimensional} with severity labels.
AutoMQM \citep{fernandes-etal-2023-devil} elicits error spans and categories through
prompting. EAPrompt \citep{lu-etal-2024-error} combines chain-of-thought with error analysis
and improves segment-level accuracy, where earlier prompting methods were weak. Rubric-MQM
\citep{kim-2025-rubric} identifies label biases and the difficulty of assessing near-perfect
translations, and improves reliability through rubric-style multidimensional prompting.
Several later systems refine this approach in different ways. MQM-APE \citep{lu-etal-2025-mqm}
adds an automatic post-editing stage that filters errors with no effect on quality.
Rubric-as-Experts \citep{xu2026rubricasexpertscasespecificmqmrubrics} allocates MQM subtype
spaces per instance and reports that broader evaluation spaces raise coverage at the cost of
false positives, while narrower spaces improve precision. GEMBA V2
\citep{junczys-dowmunt-2025-gemba}, built on the same \mGptmini backbone we use, scores each
segment ten times and aggregates the results to reduce variance. All of these systems judge a
segment with a single prompt. \facet instead separates fluency, accuracy, and consistency into
distinct passes, each given only the context its error type requires.

\paragraph{Quality estimation metrics.}
Quality estimation (QE) scores a translation from the source and the target alone, without a
reference. This is the setting the present task inherits from the QE track. Learned metrics
have led this line for years. COMET-QE \citep{chimoto-bassett-2022-comet} and TransQuest
\citep{ranasinghe-etal-2020-transquest} won the earliest WMT QE-as-a-metric subtasks, and
CometKiwi \citep{rei-etal-2022-cometkiwi} won the WMT22 QE task by jointly predicting
sentence- and word-level quality. More recent metrics add finer-grained outputs. xCOMET
\citep{guerreiro-etal-2024-xcomet} couples a learned metric with error-span prediction, and
MetricX \citep{juraska-etal-2024-metricx} scores translations with or without a reference
from a single model. These metrics are trained on human quality annotations such as the
MLQE-PE and MQM corpora. \facet stays in the reference-free QE setting but uses no trained
components, obtaining error spans, scores, and an error-free decision from prompting alone.
\begin{figure}[t]
\centering
\resizebox{\linewidth}{!}{
\begin{tikzpicture}[
    font=\sffamily,
    >=Stealth,
    node distance=0pt,
    box/.style={rounded corners=3pt, minimum height=9mm, align=center,
                inner xsep=8pt, inner ysep=5pt, font=\normalsize\bfseries},
    small/.style={font=\footnotesize},
    flow/.style={->, draw=black!65, line width=0.8pt},
    flowd/.style={->, draw=consistcol!90, line width=0.8pt, dashed},
]

  \node[box, fill=docblue!12, draw=docblue!70, text width=7.2cm] (doc)
    at (0,0) {Document};

  \node[box, fill=fluencycol!12, draw=fluencycol!70, text width=3.6cm] (flu)
    at (-4.4,-2) {\flu};
  \node[box, fill=accuracycol!15, draw=accuracycol!90, line width=1pt, text width=3.6cm] (acc)
    at (0,-2) {\acc};
  \node[box, fill=consistcol!12, draw=consistcol!70, dashed, text width=3.6cm] (con)
    at (4.4,-2) {\cst};

  \node[small, below=1pt of flu] (fli) {\textit{target}};
  \node[small, below=1pt of acc] (aci) {\textit{source\,+\,target}};
  \node[small, below=1pt of con] (coi) {\textit{whole document}};

  \node[box, fill=aggcol!12, draw=aggcol!60, text width=7.2cm] (mrg)
    at (0,-4) {Merge \& severity scoring};

  \node[box, fill=facetcol!12, draw=facetcol!85, line width=1pt, text width=3.4cm] (fc)
    at (-2.3,-6) {\facetc};
  \node[box, fill=docfacetcol!12, draw=docfacetcol!85, line width=1pt, text width=3.4cm] (ft)
    at (2.3,-6) {\facet};
  \node[small, below=1pt of fc] {\flu + \acc};
  \node[small, below=1pt of ft] {+ \cst};

  \draw[flow]  (doc.south) -- (acc.north);
  \draw[flow]  (doc.south) -| (flu.north);
  \draw[flowd] (doc.south) -| (con.north);
  \draw[flow]  (flu.south) |- (mrg.north);
  \draw[flow]  (acc.south) -- (mrg.north);
  \draw[flowd] (con.south) |- (mrg.north);
  \draw[flow]  (mrg.south) -| (fc.north);
  \draw[flow]  (mrg.south) -| (ft.north);

\end{tikzpicture}
}
\caption{\facet evaluates a translation with three independent prompting passes under
\textbf{asymmetric input}. \acc reads the source and the target, while \flu and \cst read
the target only. Predictions are merged and scored per segment. \facet\ uses all three
passes, and \facetc\ drops \cst (dashed).}
\label{fig:facet-architecture}
\end{figure}
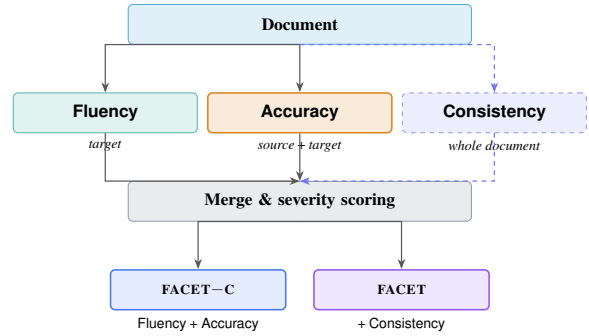

\section{System Description}
\label{sec:system}

\facet evaluates a hypothesis translation with three prompting passes that run independently
and are combined afterward. Each pass returns a list of error spans. A span carries a quoted
region of text, the erroneous substring within it, a severity, a category, and a short reason.
Figure~\ref{fig:facet-architecture} shows the pipeline. The passes differ in the input each
one receives, and this difference fixes the scope at which each pass operates. Prompts are given in \Cref{app:prompts}.

\subsection{Asymmetric Input}
The three passes do not receive the same input. \acc is the only judgment that requires both
the source and the target, since whether a translation preserves meaning cannot be decided
from the target alone. \flu and \cst require the target only. \flu asks whether a
segment is well-formed in the target language, a property of the target as a piece of writing.
\cst asks whether a recurring entity or term keeps one rendering across the document, a
property of the target across its own length. Neither question is answered by the source, and
supplying it would invite the model to import meaning-level concerns that belong to the
\acc pass. The same asymmetry fixes the scope of each pass. \flu and \acc operate on the
segment, while \cst operates on the whole document, the only scope at which the renderings
of a term can be compared against one another.

\subsection{Input Construction}
The shared task provides text in paragraph-sized units. Line breaks in the data mark
subtitle-cue boundaries rather than sentence boundaries, so we re-segment the source and the
target into sentences with a rule-based multilingual segmenter before the passes run. Newlines
are replaced with spaces in a length-preserving substitution, so the character span of each
sentence maps directly back onto the original text and all offset recovery stays in the
original coordinates. Presenting a segment as explicit sentences also gives the model a stable
structure to iterate over, which reduces its tendency to skip material in long inputs
\citep{kim-2025-longer-context}. For the \cst pass, segments sharing a document identifier
are grouped and ordered to reconstruct the full document.

\subsection{Fluency Pass}
The \flu pass reads the target segment alone and acts as a strict proofreader of the target
language. It judges spelling, grammar, agreement, word order, punctuation, orthography,
register, and terminology by the norms of the target language itself. The source is withheld
because a \flu judgment concerns the target as language, and showing the source invites
meaning-level concerns that belong to \acc.

\subsection{Accuracy Pass}
The \acc pass reads the source and the target together and judges whether the target
preserves the meaning of the source. It covers mistranslation, wrong term, addition,
hallucination, omission, and untranslated text. Source-side errors, namely omission and
defective source, are recorded on the source so that content dropped from the translation is
captured even when the target reads fluently. Hallucinated content unsupported by the source
is assigned the highest severity.

\subsection{Consistency Pass}
\label{subsec:consistency}
The \cst pass reads the whole target document and judges only internal consistency, never
correctness, which is why it needs no source. It first identifies the recurring items whose
renderings must stay stable, that is, named entities, domain terms, and fixed expressions,
and records the forms in which each appears across the document. An item with a single form
throughout is consistent by construction and is left to the \acc pass. An item with two or
more distinct forms is checked on two points, whether the forms would read as different
referents, and whether shortened or pronominal references agree in gender and number with
the form they refer back to, where the target language marks such agreement.

When an item does diverge, the error is not attached to the first or the last occurrence. The
pass decides from the document context alone which form does not fit the rest of the
document and marks every occurrence of that form, wherever it appears. Occurrences of the
form that fits are left unmarked.

\subsection{Severity and Scoring}
Every pass rates each error on one shared four-point scale, appended identically to all three
prompts. Following the severity criteria in the human evaluation guidelines provided by WMT,
scale~1 is a slight wording change with no impact on clarity or intent, scale~2 alters wording
while the message stays mostly clear, scale~3 noticeably distorts meaning so that the reader
is misled, and scale~4 is a substantial distortion that leaves the translation unfaithful. The
prompt records the raw value, and the mapping onto Minor and Major is applied afterward.
Following the correspondence in the guidelines, we cut at 3, so scales 1--2 are Minor and
scales 3--4 are Major.

Before scoring, \cst errors at scale~1 are discarded. On morphologically rich target
languages the \cst pass flags ordinary case, gender, or number inflection of a recurring
term as an inconsistency, and these false positives concentrate at scale~1. Removing them
leaves the higher-severity divergences that reflect cross-segment errors. \flu and \acc
errors at scale~1 are kept.

Because the passes are prompted and collected separately, the \flu, \acc, and \cst
predictions for a segment arrive on separate records. We merge all predictions that share a
segment before scoring rather than letting one pass overwrite another. Error spans are located
in the original text by matching the quoted region returned by the model and then the
erroneous substring within it, rather than by trusting model-reported offsets. The segment
score is the negated sum of the severities of its surviving errors across all passes, so a
segment with errors of scale 2, 1, and 3 scores $-6$ and a segment with no error scores 0. A higher score is a better translation, and a segment scores 0 exactly when no error survives.

\subsection{Deriving the Task Outputs}
One merged, scored prediction yields all three shared-task outputs. For error span detection, each surviving error contributes its character offsets, its Minor or Major label under the cut, and its category. \cst errors whose referents span multiple non-contiguous positions may lack a single offset, and these are omitted from the span output. For quality scoring, the segment score above is used directly. For error-free detection, a segment is labeled error-free when its score is 0, that is, when no error survives the merge.
\begin{table*}[t]
\centering
\setlength{\tabcolsep}{3.5pt}
\renewcommand{\arraystretch}{1.1}
\small
\begin{tabular}{lccccccccccccccccccccc|c}
\toprule
System & \rotatebox{90}{cs-de} & \rotatebox{90}{en-de} & \rotatebox{90}{cs-vi} & \rotatebox{90}{en-is} & \rotatebox{90}{en-et} & \rotatebox{90}{zh-ja} & \rotatebox{90}{en-se} & \rotatebox{90}{en-zh} & \rotatebox{90}{en-th} & \rotatebox{90}{en-cs} & \rotatebox{90}{en-kk} & \rotatebox{90}{en-ar} & \rotatebox{90}{en-ru} & \rotatebox{90}{cs-uk} & \rotatebox{90}{en-lij} & \rotatebox{90}{en-uk} & \rotatebox{90}{en-lld} & \rotatebox{90}{en-hy} & \rotatebox{90}{en-be} & \rotatebox{90}{en-id} & \rotatebox{90}{en-zh-TW} & \rotatebox{90}{\textbf{avg}} \\
\midrule
\emph{Human (postediting)} & \cellcolor{rankblue!55}1 & \cellcolor{rankblue!53}2 & \cellcolor{rankblue!55}1 & \cellcolor{rankblue!0}-- & \cellcolor{rankblue!0}-- & \cellcolor{rankblue!55}1 & \cellcolor{rankblue!0}-- & \cellcolor{rankblue!0}-- & \cellcolor{rankblue!0}-- & \cellcolor{rankblue!55}1 & \cellcolor{rankblue!0}-- & \cellcolor{rankblue!53}2 & \cellcolor{rankblue!0}-- & \cellcolor{rankblue!0}-- & \cellcolor{rankblue!0}-- & \cellcolor{rankblue!0}-- & \cellcolor{rankblue!0}-- & \cellcolor{rankblue!0}-- & \cellcolor{rankblue!0}-- & \cellcolor{rankblue!0}-- & \cellcolor{rankblue!0}-- & \textbf{1.3} \\
\emph{Human (from scratch)} & \cellcolor{rankblue!29}15 & \cellcolor{rankblue!0}-- & \cellcolor{rankblue!29}15 & \cellcolor{rankblue!42}8 & \cellcolor{rankblue!40}9 & \cellcolor{rankblue!0}-- & \cellcolor{rankblue!37}11 & \cellcolor{rankblue!0}-- & \cellcolor{rankblue!0}-- & \cellcolor{rankblue!0}-- & \cellcolor{rankblue!0}-- & \cellcolor{rankblue!0}-- & \cellcolor{rankblue!0}-- & \cellcolor{rankblue!0}-- & \cellcolor{rankblue!20}20 & \cellcolor{rankblue!0}-- & \cellcolor{rankblue!29}15 & \cellcolor{rankblue!0}-- & \cellcolor{rankblue!0}-- & \cellcolor{rankblue!0}-- & \cellcolor{rankblue!0}-- & \textbf{13.3} \\
\midrule
Wayfinder & \cellcolor{rankblue!48}5 & \cellcolor{rankblue!51}3 & \cellcolor{rankblue!48}5 & \cellcolor{rankblue!55}1 & \cellcolor{rankblue!53}2 & \cellcolor{rankblue!48}5 & \cellcolor{rankblue!48}5 & \cellcolor{rankblue!53}2 & \cellcolor{rankblue!51}3 & \cellcolor{rankblue!38}10 & \cellcolor{rankblue!55}1 & \cellcolor{rankblue!55}1 & \cellcolor{rankblue!51}3 & \cellcolor{rankblue!53}2 & \cellcolor{rankblue!53}2 & \cellcolor{rankblue!55}1 & \cellcolor{rankblue!37}11 & \cellcolor{rankblue!48}5 & \cellcolor{rankblue!51}3 & \cellcolor{rankblue!20}20 & \cellcolor{rankblue!50}4 & \textbf{4.5} \\
Lumen & \cellcolor{rankblue!40}9 & \cellcolor{rankblue!44}7 & \cellcolor{rankblue!51}3 & \cellcolor{rankblue!53}2 & \cellcolor{rankblue!50}4 & \cellcolor{rankblue!42}8 & \cellcolor{rankblue!51}3 & \cellcolor{rankblue!50}4 & \cellcolor{rankblue!55}1 & \cellcolor{rankblue!48}5 & \cellcolor{rankblue!50}4 & \cellcolor{rankblue!40}9 & \cellcolor{rankblue!46}6 & \cellcolor{rankblue!55}1 & \cellcolor{rankblue!55}1 & \cellcolor{rankblue!50}4 & \cellcolor{rankblue!50}4 & \cellcolor{rankblue!53}2 & \cellcolor{rankblue!55}1 & \cellcolor{rankblue!37}11 & \cellcolor{rankblue!46}6 & \textbf{4.5} \\
GPT 5.5 & \cellcolor{rankblue!53}2 & \cellcolor{rankblue!46}6 & \cellcolor{rankblue!53}2 & \cellcolor{rankblue!51}3 & \cellcolor{rankblue!55}1 & \cellcolor{rankblue!51}3 & \cellcolor{rankblue!44}7 & \cellcolor{rankblue!38}10 & \cellcolor{rankblue!53}2 & \cellcolor{rankblue!50}4 & \cellcolor{rankblue!46}6 & \cellcolor{rankblue!51}3 & \cellcolor{rankblue!42}8 & \cellcolor{rankblue!51}3 & \cellcolor{rankblue!37}11 & \cellcolor{rankblue!46}6 & \cellcolor{rankblue!51}3 & \cellcolor{rankblue!51}3 & \cellcolor{rankblue!53}2 & \cellcolor{rankblue!51}3 & \cellcolor{rankblue!40}9 & \textbf{4.6} \\
UvA-MT & \cellcolor{rankblue!51}3 & \cellcolor{rankblue!50}4 & \cellcolor{rankblue!42}8 & \cellcolor{rankblue!44}7 & \cellcolor{rankblue!48}5 & \cellcolor{rankblue!53}2 & \cellcolor{rankblue!53}2 & \cellcolor{rankblue!55}1 & \cellcolor{rankblue!50}4 & \cellcolor{rankblue!42}8 & \cellcolor{rankblue!51}3 & \cellcolor{rankblue!44}7 & \cellcolor{rankblue!48}5 & \cellcolor{rankblue!44}7 & \cellcolor{rankblue!51}3 & \cellcolor{rankblue!48}5 & \cellcolor{rankblue!44}7 & \cellcolor{rankblue!50}4 & \cellcolor{rankblue!48}5 & \cellcolor{rankblue!40}9 & \cellcolor{rankblue!48}5 & \textbf{5.0} \\
TRIVE & \cellcolor{rankblue!44}7 & \cellcolor{rankblue!48}5 & \cellcolor{rankblue!46}6 & \cellcolor{rankblue!48}5 & \cellcolor{rankblue!51}3 & \cellcolor{rankblue!50}4 & \cellcolor{rankblue!42}8 & \cellcolor{rankblue!48}5 & \cellcolor{rankblue!44}7 & \cellcolor{rankblue!40}9 & \cellcolor{rankblue!53}2 & \cellcolor{rankblue!48}5 & \cellcolor{rankblue!50}4 & \cellcolor{rankblue!50}4 & \cellcolor{rankblue!44}7 & \cellcolor{rankblue!53}2 & \cellcolor{rankblue!42}8 & \cellcolor{rankblue!55}1 & \cellcolor{rankblue!50}4 & \cellcolor{rankblue!29}15 & \cellcolor{rankblue!51}3 & \textbf{5.4} \\
Cohere CAT+ & \cellcolor{rankblue!50}4 & \cellcolor{rankblue!55}1 & \cellcolor{rankblue!50}4 & \cellcolor{rankblue!31}14 & \cellcolor{rankblue!35}12 & \cellcolor{rankblue!38}10 & \cellcolor{rankblue!55}1 & \cellcolor{rankblue!51}3 & \cellcolor{rankblue!40}9 & \cellcolor{rankblue!53}2 & \cellcolor{rankblue!28}16 & \cellcolor{rankblue!42}8 & \cellcolor{rankblue!53}2 & \cellcolor{rankblue!48}5 & \cellcolor{rankblue!48}5 & \cellcolor{rankblue!51}3 & \cellcolor{rankblue!53}2 & \cellcolor{rankblue!24}18 & \cellcolor{rankblue!33}13 & \cellcolor{rankblue!55}1 & \cellcolor{rankblue!53}2 & \textbf{6.4} \\
Dubformer & \cellcolor{rankblue!42}8 & \cellcolor{rankblue!38}10 & \cellcolor{rankblue!44}7 & \cellcolor{rankblue!28}16 & \cellcolor{rankblue!33}13 & \cellcolor{rankblue!46}6 & \cellcolor{rankblue!38}10 & \cellcolor{rankblue!46}6 & \cellcolor{rankblue!35}12 & \cellcolor{rankblue!37}11 & \cellcolor{rankblue!42}8 & \cellcolor{rankblue!31}14 & \cellcolor{rankblue!44}7 & \cellcolor{rankblue!46}6 & \cellcolor{rankblue!38}10 & \cellcolor{rankblue!42}8 & \cellcolor{rankblue!31}14 & \cellcolor{rankblue!33}13 & \cellcolor{rankblue!40}9 & \cellcolor{rankblue!48}5 & \cellcolor{rankblue!44}7 & \textbf{9.5} \\
Gemini 3.1 Pro & \cellcolor{rankblue!33}13 & \cellcolor{rankblue!26}17 & \cellcolor{rankblue!33}13 & \cellcolor{rankblue!37}11 & \cellcolor{rankblue!44}7 & \cellcolor{rankblue!40}9 & \cellcolor{rankblue!35}12 & \cellcolor{rankblue!40}9 & \cellcolor{rankblue!48}5 & \cellcolor{rankblue!31}14 & \cellcolor{rankblue!44}7 & \cellcolor{rankblue!38}10 & \cellcolor{rankblue!37}11 & \cellcolor{rankblue!35}12 & \cellcolor{rankblue!42}8 & \cellcolor{rankblue!37}11 & \cellcolor{rankblue!46}6 & \cellcolor{rankblue!42}8 & \cellcolor{rankblue!44}7 & \cellcolor{rankblue!15}23 & \cellcolor{rankblue!42}8 & \textbf{10.5} \\
MakotoAI & \cellcolor{rankblue!38}10 & \cellcolor{rankblue!28}16 & \cellcolor{rankblue!28}16 & \cellcolor{rankblue!46}6 & \cellcolor{rankblue!46}6 & \cellcolor{rankblue!26}17 & \cellcolor{rankblue!50}4 & \cellcolor{rankblue!28}16 & \cellcolor{rankblue!31}14 & \cellcolor{rankblue!46}6 & \cellcolor{rankblue!37}11 & \cellcolor{rankblue!50}4 & \cellcolor{rankblue!28}16 & \cellcolor{rankblue!40}9 & \cellcolor{rankblue!29}15 & \cellcolor{rankblue!44}7 & \cellcolor{rankblue!35}12 & \cellcolor{rankblue!40}9 & \cellcolor{rankblue!35}12 & \cellcolor{rankblue!26}17 & \cellcolor{rankblue!29}15 & \textbf{11.3} \\
DeepSeek V4 Pro & \cellcolor{rankblue!37}11 & \cellcolor{rankblue!31}14 & \cellcolor{rankblue!26}17 & \cellcolor{rankblue!38}10 & \cellcolor{rankblue!31}14 & \cellcolor{rankblue!31}14 & \cellcolor{rankblue!46}6 & \cellcolor{rankblue!35}12 & \cellcolor{rankblue!28}16 & \cellcolor{rankblue!35}12 & \cellcolor{rankblue!35}12 & \cellcolor{rankblue!37}11 & \cellcolor{rankblue!29}15 & \cellcolor{rankblue!37}11 & \cellcolor{rankblue!35}12 & \cellcolor{rankblue!40}9 & \cellcolor{rankblue!48}5 & \cellcolor{rankblue!44}7 & \cellcolor{rankblue!42}8 & \cellcolor{rankblue!35}12 & \cellcolor{rankblue!37}11 & \textbf{11.4} \\
Qingqiu-MT-9B & \cellcolor{rankblue!28}16 & \cellcolor{rankblue!29}15 & \cellcolor{rankblue!37}11 & \cellcolor{rankblue!24}18 & \cellcolor{rankblue!22}19 & \cellcolor{rankblue!35}12 & \cellcolor{rankblue!24}18 & \cellcolor{rankblue!44}7 & \cellcolor{rankblue!37}11 & \cellcolor{rankblue!28}16 & \cellcolor{rankblue!38}10 & \cellcolor{rankblue!26}17 & \cellcolor{rankblue!33}13 & \cellcolor{rankblue!31}14 & \cellcolor{rankblue!50}4 & \cellcolor{rankblue!29}15 & \cellcolor{rankblue!38}10 & \cellcolor{rankblue!37}11 & \cellcolor{rankblue!31}14 & \cellcolor{rankblue!31}14 & \cellcolor{rankblue!35}12 & \textbf{13.2} \\
Gemma 4 - 31B & \cellcolor{rankblue!31}14 & \cellcolor{rankblue!24}18 & \cellcolor{rankblue!40}9 & \cellcolor{rankblue!17}22 & \cellcolor{rankblue!13}24 & \cellcolor{rankblue!37}11 & \cellcolor{rankblue!18}21 & \cellcolor{rankblue!33}13 & \cellcolor{rankblue!38}10 & \cellcolor{rankblue!18}21 & \cellcolor{rankblue!29}15 & \cellcolor{rankblue!35}12 & \cellcolor{rankblue!35}12 & \cellcolor{rankblue!33}13 & \cellcolor{rankblue!33}13 & \cellcolor{rankblue!31}14 & \cellcolor{rankblue!22}19 & \cellcolor{rankblue!22}19 & \cellcolor{rankblue!18}21 & \cellcolor{rankblue!46}6 & \cellcolor{rankblue!31}14 & \textbf{15.3} \\
Command A+ & \cellcolor{rankblue!35}12 & \cellcolor{rankblue!40}9 & \cellcolor{rankblue!22}19 & \cellcolor{rankblue!18}21 & \cellcolor{rankblue!24}18 & \cellcolor{rankblue!20}20 & \cellcolor{rankblue!7}27 & \cellcolor{rankblue!31}14 & \cellcolor{rankblue!22}19 & \cellcolor{rankblue!44}7 & \cellcolor{rankblue!18}21 & \cellcolor{rankblue!24}18 & \cellcolor{rankblue!24}18 & \cellcolor{rankblue!29}15 & \cellcolor{rankblue!28}16 & \cellcolor{rankblue!28}16 & \cellcolor{rankblue!26}17 & \cellcolor{rankblue!5}28 & \cellcolor{rankblue!26}17 & \cellcolor{rankblue!44}7 & \cellcolor{rankblue!28}16 & \textbf{16.9} \\
Mistral Medium 3.5 & \cellcolor{rankblue!26}17 & \cellcolor{rankblue!33}13 & \cellcolor{rankblue!24}18 & \cellcolor{rankblue!15}23 & \cellcolor{rankblue!17}22 & \cellcolor{rankblue!18}21 & \cellcolor{rankblue!28}16 & \cellcolor{rankblue!29}15 & \cellcolor{rankblue!15}23 & \cellcolor{rankblue!29}15 & \cellcolor{rankblue!15}23 & \cellcolor{rankblue!17}22 & \cellcolor{rankblue!22}19 & \cellcolor{rankblue!28}16 & \cellcolor{rankblue!46}6 & \cellcolor{rankblue!26}17 & \cellcolor{rankblue!40}9 & \cellcolor{rankblue!26}17 & \cellcolor{rankblue!20}20 & \cellcolor{rankblue!33}13 & \cellcolor{rankblue!24}18 & \textbf{17.3} \\
GPT OSS 120B & \cellcolor{rankblue!20}20 & \cellcolor{rankblue!22}19 & \cellcolor{rankblue!11}25 & \cellcolor{rankblue!20}20 & \cellcolor{rankblue!20}20 & \cellcolor{rankblue!24}18 & \cellcolor{rankblue!29}15 & \cellcolor{rankblue!24}18 & \cellcolor{rankblue!11}25 & \cellcolor{rankblue!24}18 & \cellcolor{rankblue!20}20 & \cellcolor{rankblue!20}20 & \cellcolor{rankblue!17}22 & \cellcolor{rankblue!18}21 & \cellcolor{rankblue!40}9 & \cellcolor{rankblue!22}19 & \cellcolor{rankblue!55}1 & \cellcolor{rankblue!15}23 & \cellcolor{rankblue!17}22 & \cellcolor{rankblue!24}18 & \cellcolor{rankblue!33}13 & \textbf{18.4} \\
SalamandraTA & \cellcolor{rankblue!18}21 & \cellcolor{rankblue!20}20 & \cellcolor{rankblue!9}26 & \cellcolor{rankblue!33}13 & \cellcolor{rankblue!26}17 & \cellcolor{rankblue!7}27 & \cellcolor{rankblue!26}17 & \cellcolor{rankblue!11}25 & \cellcolor{rankblue!17}22 & \cellcolor{rankblue!22}19 & \cellcolor{rankblue!22}19 & \cellcolor{rankblue!11}25 & \cellcolor{rankblue!13}24 & \cellcolor{rankblue!26}17 & \cellcolor{rankblue!15}23 & \cellcolor{rankblue!18}21 & \cellcolor{rankblue!13}24 & \cellcolor{rankblue!29}15 & \cellcolor{rankblue!15}23 & \cellcolor{rankblue!9}26 & \cellcolor{rankblue!9}26 & \textbf{21.4} \\
Qwen 3.6 - 27b & \cellcolor{rankblue!15}23 & \cellcolor{rankblue!15}23 & \cellcolor{rankblue!17}22 & \cellcolor{rankblue!11}25 & \cellcolor{rankblue!15}23 & \cellcolor{rankblue!11}25 & \cellcolor{rankblue!31}14 & \cellcolor{rankblue!17}22 & \cellcolor{rankblue!13}24 & \cellcolor{rankblue!11}25 & \cellcolor{rankblue!13}24 & \cellcolor{rankblue!15}23 & \cellcolor{rankblue!11}25 & \cellcolor{rankblue!20}20 & \cellcolor{rankblue!22}19 & \cellcolor{rankblue!11}25 & \cellcolor{rankblue!20}20 & \cellcolor{rankblue!18}21 & \cellcolor{rankblue!13}24 & \cellcolor{rankblue!13}24 & \cellcolor{rankblue!20}20 & \textbf{22.4} \\
Gemma 4 - 4B & \cellcolor{rankblue!11}25 & \cellcolor{rankblue!17}22 & \cellcolor{rankblue!18}21 & \cellcolor{rankblue!2}30 & \cellcolor{rankblue!5}28 & \cellcolor{rankblue!22}19 & \cellcolor{rankblue!9}26 & \cellcolor{rankblue!15}23 & \cellcolor{rankblue!24}18 & \cellcolor{rankblue!5}28 & \cellcolor{rankblue!17}22 & \cellcolor{rankblue!22}19 & \cellcolor{rankblue!15}23 & \cellcolor{rankblue!15}23 & \cellcolor{rankblue!13}24 & \cellcolor{rankblue!20}20 & \cellcolor{rankblue!11}25 & \cellcolor{rankblue!13}24 & \cellcolor{rankblue!4}29 & \cellcolor{rankblue!22}19 & \cellcolor{rankblue!18}21 & \textbf{23.3} \\
Ministral 3 - 14B & \cellcolor{rankblue!17}22 & \cellcolor{rankblue!18}21 & \cellcolor{rankblue!13}24 & \cellcolor{rankblue!9}26 & \cellcolor{rankblue!11}25 & \cellcolor{rankblue!9}26 & \cellcolor{rankblue!15}23 & \cellcolor{rankblue!7}27 & \cellcolor{rankblue!7}27 & \cellcolor{rankblue!13}24 & \cellcolor{rankblue!7}27 & \cellcolor{rankblue!7}27 & \cellcolor{rankblue!9}26 & \cellcolor{rankblue!17}22 & \cellcolor{rankblue!17}22 & \cellcolor{rankblue!13}24 & \cellcolor{rankblue!28}16 & \cellcolor{rankblue!20}20 & \cellcolor{rankblue!11}25 & \cellcolor{rankblue!7}27 & \cellcolor{rankblue!11}25 & \textbf{24.1} \\
Tower 9B & \cellcolor{rankblue!13}24 & \cellcolor{rankblue!11}25 & \cellcolor{rankblue!0}31 & \cellcolor{rankblue!26}17 & \cellcolor{rankblue!2}30 & \cellcolor{rankblue!13}24 & \cellcolor{rankblue!11}25 & \cellcolor{rankblue!5}28 & \cellcolor{rankblue!2}30 & \cellcolor{rankblue!17}22 & \cellcolor{rankblue!5}28 & \cellcolor{rankblue!2}30 & \cellcolor{rankblue!5}28 & \cellcolor{rankblue!22}19 & \cellcolor{rankblue!24}18 & \cellcolor{rankblue!17}22 & \cellcolor{rankblue!17}22 & \cellcolor{rankblue!7}27 & \cellcolor{rankblue!2}30 & \cellcolor{rankblue!2}30 & \cellcolor{rankblue!15}23 & \textbf{25.4} \\
Qwen 3.5 - 9b & \cellcolor{rankblue!5}28 & \cellcolor{rankblue!5}28 & \cellcolor{rankblue!5}28 & \cellcolor{rankblue!5}28 & \cellcolor{rankblue!9}26 & \cellcolor{rankblue!5}28 & \cellcolor{rankblue!20}20 & \cellcolor{rankblue!18}21 & \cellcolor{rankblue!9}26 & \cellcolor{rankblue!2}30 & \cellcolor{rankblue!9}26 & \cellcolor{rankblue!4}29 & \cellcolor{rankblue!7}27 & \cellcolor{rankblue!9}26 & \cellcolor{rankblue!7}27 & \cellcolor{rankblue!7}27 & \cellcolor{rankblue!15}23 & \cellcolor{rankblue!11}25 & \cellcolor{rankblue!9}26 & \cellcolor{rankblue!4}29 & \cellcolor{rankblue!17}22 & \textbf{26.2} \\
GPT OSS 20B & \cellcolor{rankblue!4}29 & \cellcolor{rankblue!7}27 & \cellcolor{rankblue!4}29 & \cellcolor{rankblue!7}27 & \cellcolor{rankblue!7}27 & \cellcolor{rankblue!4}29 & \cellcolor{rankblue!22}19 & \cellcolor{rankblue!9}26 & \cellcolor{rankblue!4}29 & \cellcolor{rankblue!7}27 & \cellcolor{rankblue!11}25 & \cellcolor{rankblue!9}26 & \cellcolor{rankblue!2}30 & \cellcolor{rankblue!5}28 & \cellcolor{rankblue!9}26 & \cellcolor{rankblue!5}28 & \cellcolor{rankblue!18}21 & \cellcolor{rankblue!9}26 & \cellcolor{rankblue!7}27 & \cellcolor{rankblue!5}28 & \cellcolor{rankblue!13}24 & \textbf{26.6} \\
GLM 4 - 9B & \cellcolor{rankblue!0}31 & \cellcolor{rankblue!4}29 & \cellcolor{rankblue!2}30 & \cellcolor{rankblue!0}31 & \cellcolor{rankblue!4}29 & \cellcolor{rankblue!0}31 & \cellcolor{rankblue!13}24 & \cellcolor{rankblue!13}24 & \cellcolor{rankblue!0}31 & \cellcolor{rankblue!4}29 & \cellcolor{rankblue!4}29 & \cellcolor{rankblue!5}28 & \cellcolor{rankblue!0}31 & \cellcolor{rankblue!4}29 & \cellcolor{rankblue!4}29 & \cellcolor{rankblue!2}30 & \cellcolor{rankblue!7}27 & \cellcolor{rankblue!17}22 & \cellcolor{rankblue!0}31 & \cellcolor{rankblue!0}31 & \cellcolor{rankblue!7}27 & \textbf{28.7} \\
\bottomrule
\end{tabular}
\caption{\facet system rankings across the 21 language pairs. Each cell is the rank of a
system within that language pair (1 = best), darker is better, and \texttt{--} marks a pair
the system did not enter. Systems below the rule are ordered by mean rank. \emph{Human
(postediting)} attains the best mean rank, while \emph{Human (from scratch)} sits mid-table.}
\label{tab:system-ranking}
\end{table*}

\section{Experimental Setup}
\label{sec:setup}
\paragraph{Model.}
All three passes use a single model, \mGptmini (\texttt{gpt-4.1-mini-2025-04-14}), served
through the OpenAI API and accessed only through prompting with no fine-tuning. It is not an
ensemble and is not based on open weights, and its parameter count is not publicly disclosed.
We decode deterministically at temperature 0, so the judgments of \facet are reproducible.
The model is non-reasoning, so its output is the answer itself with no intermediate reasoning
trace. The maximum output length is set per request in proportion to the input. For the
segment-level \flu and \acc passes we allow $1.5\times$ the input length plus a floor
of 1{,}000 tokens, capped at 8{,}000. For the document-level \cst pass, whose output
must enumerate every tracked item, we allow $2\times$ the input plus a floor of 2{,}000
tokens, capped at 16{,}000. Requests are issued through the batch interface, one per segment for \flu, one per segment for \acc, and one per document for \cst.

\paragraph{Data.}
We evaluate on the WMT26 test set, which spans 21 language pairs and multiple domains, from
official speech-sourced material to challenge sets. Each item pairs a source with the outputs
of many translation systems, and \facet scores every system output independently, yielding
several hundred thousand system-segment pairs per pass. The human annotations and the
official leaderboard were not released at the time of writing, so we do not report
correlation with gold labels.

\paragraph{Speech transcripts.}
A large share of the data is speech-sourced. In pilot runs the \flu and \acc passes flagged filler words and hesitations, such as \textit{um} and \textit{uh} and their target-language equivalents, as errors, and treated their differing renderings across languages as mistakes. These are transcription artifacts, not translation errors. For speech-based items we therefore append a rule instructing the \flu and \acc passes to ignore fillers, hesitations, and repeated interjections. The \cst pass is left unchanged, since fillers are not recurring entities or terms.

\paragraph{Submissions.}
Our primary submission, \facet, uses all three passes with the scale-1 filtering of \cst errors described in Section~\ref{sec:system}. To isolate what the document-level \cst pass contributes under segment-level scoring, we also submit \facetc, an ablation that omits \cst and scores from \flu and \acc alone.

\section{Submission Analysis}
\label{sec:analysis}
Without gold labels, we characterize the predictions of \facet on their own terms, through the system ranking they induce, the effect of the \cst pass, and one worked example.

\paragraph{System ranking.}
For each language pair we rank the systems by their mean \facet score and report the per-pair
ranks in Table~\ref{tab:system-ranking}. Because languages differ in how many errors the
passes surface, ranking within a language pair rather than pooling across pairs keeps
language difficulty from being confounded with system quality. The resulting order follows
the pattern one would expect from the systems involved. Strong general-purpose systems such
as Wayfinder, Lumen, and GPT~5.5 rank consistently near the top, and small open models such
as GLM~4-9B, GPT~OSS~20B, and Qwen~3.5-9b rank consistently near the bottom.

The clearest pattern is in the human entries. \emph{Human (postediting)} attains the best
mean rank of any system at 1.3, placing first or second in every pair it enters, while
\emph{Human (from scratch)} sits mid-table at 13.3. One reading is that \facet rewards
polished translation over first-draft translation and does not simply favor human text.
Another is that a reference-free LLM judge is more tolerant of translations that stay close
to the source structure, as post-edited MT does, and penalizes the freer choices of
translation from scratch. The two readings cannot be separated without human judgments, and
we leave the question to the official evaluation.

\begin{figure*}[t]
\centering\small
\begin{tcolorbox}[colback=white,colframe=black!45,boxrule=0.5pt,arc=2pt,left=7pt,right=7pt,top=6pt,bottom=6pt]
A full Czech--German document evaluated by \facet. Every detected error is highlighted: \hlF{\flu}, \hlA{\acc}, \hlC{\cst}.
\medskip

\textbf{Segment 1 --- source.} <p>Pilníkárna v Křižánkách na Žďársku se stala kulturní památkou</p>

\textbf{Segment 1 --- target.} <p>Die \hlC{Bürstenfabrik} in Křižánky \hlA{im Žďárský-Krkonose} ist zu einem Kulturdenkmal geworden</p> 
\medskip

\textbf{Segment 2 --- source.} <p>Pilníkárna v Křižánkách na Žďársku, která je součástí historického areálu Kupcova vodního mlýna a pily, se stala kulturní památkou. Navzdory úpravám je autentickým svědectvím zaniklého řemesla výroby železných pilníků. V tiskové zprávě to za telčské pracoviště Národního památkového ústavu uvedla Ilona Ampapová. Na Vysočině je téměř 3000 nemovitých kulturních památek, z toho 16 národních kulturních památek. Křižánky jsou velkou vesnickou památkovou rezervací. Památkově chráněná byla zatím jen pila a mlýn zmiňovaný poprvé v roce 1713. "Celý areál, včetně pilníkárny, je doložen již na mapě stabilního katastru z roku 1839," uvedla Ampapová. Pilníkárna vznikla v roce 1905 a její současná podoba odpovídá stavebním úpravám spojeným se zavedením pilníkářského řemesla. "Navzdory dílčím proměnám ve 20. století se zde dochovala řada autentických konstrukcí a prvků, které svědčí o původní funkci objektu a jeho historické hodnotě," uvedla Ampapová. Provoz pily i pilníkářská výroba v areálu skončila v roce 1947. Křižánky jsou největší vesnickou památkovou rezervací na Vysočině. V roce 2012 se staly Vesnicí roku Kraje Vysočina. Pilníkářskou tradici v obci připomíná od roku 2011 naučná stezka.</p>

\textbf{Segment 2 --- target.} Die \hlC{Schleiferei} in Křižánky am Žďárský-Berg, die Teil des historischen Geländes des Kupcova-Wasser-Mühlen- und Sägewerks ist, wurde zu einem Kulturdenkmal. Trotz der Anpassungen ist sie ein authentisches Zeugnis des verschollenen Handwerks der Herstellung von \hlA{Eisen-Schleifscheiben}. In einer Pressemitteilung gab Ilona Ampapová \hlF{für das Telčer Standort} des Nationalen Denkmalinstituts dies an. In der Region Vysočina gibt es fast 3000 unbewegliche Kulturdenkmäler, davon 16 nationale Kulturdenkmäler. Křižánky sind ein großes Dorfdenkmalsreservat. Bisher waren nur \hlF{der Sägewerk und der Mühle}, die erstmals im Jahr 1713 erwähnt wurden, \hlF{unter Denkmalschutz gestellt. „}Das gesamte Gelände, einschließlich der \hlA{Sägewerkshalle}, ist bereits auf der Karte des festen Katasters aus dem Jahr 1839 belegt“, erklärte Ampapová. Die Schleiferei wurde im Jahr 1905 gegründet, und ihre heutige Gestalt entspricht den baulichen Anpassungen, die mit der Einführung des Schleifereihandwerks verbunden waren. \hlF{„}Trotz der teilweisen Veränderungen im 20. Jahrhundert sind hier zahlreiche authentische Konstruktionen und Elemente erhalten geblieben, die von der ursprünglichen Funktion des Objekts und seinem historischen Wert \hlF{zeugen“, erklärte} Ampapová. Der Betrieb des Sägewerks sowie die \hlA{Sägewerkzeugherstellung} auf dem Gelände endeten im Jahr 1947. Křižánky sind die größte dörfliche Denkmalschutzzone in der Region Vysočina. Im Jahr 2012 wurden sie zum Dorf des Jahres der Region Vysočina. Seit 2011 erinnert ein Lehrpfad an die \hlA{Tradition der Pilger}.
\end{tcolorbox}
\caption{A full document evaluated by \facet, with all detected errors highlighted.
\flu and \acc act within a segment, from the target alone and from the source--target
pair. \cst acts across the document. The facility named \emph{B\"urstenfabrik} in
Segment~1 becomes \emph{Schleiferei} in Segment~2, a divergence that no single-segment
judgment can detect.}
\label{fig:example}
\end{figure*}

\paragraph{Effect of the consistency pass.}
Comparing the full \facet ranking with the two-pass ablation \facetc, the \cst pass
changes the score of 2{,}726 of 26{,}885 items (10.1\%), but the system ranking is almost
untouched. No system moves by more than 0.2 in mean rank and no two systems swap order. The effect on error-free detection is smaller still. Only 0.7\% of segments receive a different error-free label under \facet than under \facetc.
Part of this is structural, since \cst errors are document-level and rare relative to
segment-level \flu and \acc errors. The pass also recovers fewer errors than its
design allows. It is the only pass that takes a whole document as input, and the model
struggles to maintain a reliable account of recurring items across such long contexts. This
shows in the low number of \cst errors it recovers and in the share of flagged spans
it cannot localize. The divergences it does catch are cross-segment errors that \flu and
\acc cannot see, such as one entity rendered under several names within a document. We
take the weak effect as a limitation of long-context reliability in the current model rather than as evidence that document-level \cst does not matter \citep{kim-2025-longer-context}.

\paragraph{A qualitative example.}
Figure~\ref{fig:example} shows the three passes on one document. \flu and \acc stay within
a segment. \flu flags target-language faults such as the gender agreement in \emph{f\"ur
das Tel\v{c}er Standort} and the misplaced quotation marks, and \acc flags meaning errors
against the source, among them the region name \emph{\v{Z}\v{d}\'arsk\'y-Krkonose}, where
the source region \emph{\v{Z}\v{d}\'arsko} is rendered as a mountain range, and
\emph{Tradition der Pilger}, which turns a file-making tradition into a tradition of
pilgrims. The \cst pass catches an error of a different kind. The same facility is called
\emph{B\"urstenfabrik} (`brush factory') in the title segment and \emph{Schleiferei}
(`grinding shop') in the body segment, two names for one referent that are each locally
acceptable and conflict only when the segments are read together. This is the error class
the pass is designed for, and it cannot be recovered by scoring segments in isolation. The
example also shows a miss. Both \emph{B\"urstenfabrik} and \emph{Schleiferei} mistranslate
the source \emph{piln\'ik\'arna} (`file workshop'), and the \acc pass flags neither, so the
document is penalized for the divergence but not for the term itself.
\section{Conclusion}
We presented \facet, a reference-free translation quality evaluator built by decomposition
rather than by a single prompt or a trained metric. Error types need different context.
\acc is judged from the source and the target together, while \flu and \cst
are judged from the target alone, at the sentence and the document scope respectively. A
single fixed model is prompted three times, and its predictions are merged and scored, with
no trained components anywhere in the pipeline.

Without gold labels, we characterized the predictions of \facet on their own terms. Its
per-language-pair system rankings follow the expected order of the participating systems and
place post-edited human translation at the top. Whether the mid-table position of
translation from scratch reflects quality or a preference for source-close output remains
open until human judgments are released. The document-level \cst pass changes a tenth
of segment scores but leaves the system ranking almost unchanged, a result our paired
submissions were designed to expose. We attribute this to the limited reliability with which
current models track recurring items across long inputs, and to the mismatch between a
document-level signal and segment-level scoring. Stronger long-context models, or a scoring
scheme that credits document-level \cst directly, would address these two limits.

\section*{Acknowledgment}
This work was supported by the G-LAMP Program of the National Research Foundation of Korea (NRF) grant funded by the Ministry of Education (No. RS-2025-25441317); the Ministry of Science and ICT (MSIT) and the National IT Industry Promotion Agency (NIPA) through the Advanced GPU Utilization Support Program (02-26-01-0499); the NRF grant funded by the MSIT (RS-2025-16071992).
This work was supported by Korea Internet \& Security Agency(KISA) grant funded by the Korea government(PIPC) (No.RS-2026-25526342, Development of Technologies for Preventing Sensitive Information Inference and Risk Assessment in Foundation Model Operations).
This research was supported by Culture, Sports and Tourism R\&D Program through the Korea Creative Content Agency grant funded by the Ministry of Culture, Sports and Tourism in 2026 (Project Name: Develop AI agent technology to connect knowledge through public cultural facility-based discussion and communication, Project Number: RS-2026-25520645).

\bibliography{custom,anthology-2,arxiv/arxiv}

@inproceedings{fernandes-etal-2023-devil,
    title = "The Devil Is in the Errors: Leveraging Large Language Models for Fine-grained Machine Translation Evaluation",
    author = "Fernandes, Patrick  and
      Deutsch, Daniel  and
      Finkelstein, Mara  and
      Riley, Parker  and
      Martins, Andr{\'e}  and
      Neubig, Graham  and
      Garg, Ankush  and
      Clark, Jonathan  and
      Freitag, Markus  and
      Firat, Orhan",
    editor = "Koehn, Philipp  and
      Haddow, Barry  and
      Kocmi, Tom  and
      Monz, Christof",
    booktitle = "Proceedings of the Eighth Conference on Machine Translation",
    month = dec,
    year = "2023",
    address = "Singapore",
    publisher = "Association for Computational Linguistics",
    url = "https://aclanthology.org/2023.wmt-1.100/",
    doi = "10.18653/v1/2023.wmt-1.100",
    pages = "1066--1083"
}

@article{guerreiro-etal-2024-xcomet,
    title = "x{COMET}: Transparent Machine Translation Evaluation through Fine-grained Error Detection",
    author = "Guerreiro, Nuno M.  and
      Rei, Ricardo  and
      van Stigt, Daan  and
      Coheur, Luisa  and
      Colombo, Pierre  and
      Martins, Andr{\'e} F. T.",
    journal = "Transactions of the Association for Computational Linguistics",
    volume = "12",
    year = "2024",
    address = "Cambridge, MA",
    publisher = "MIT Press",
    url = "https://aclanthology.org/2024.tacl-1.54/",
    doi = "10.1162/tacl_a_00683",
    pages = "979--995"
}

@inproceedings{junczys-dowmunt-2025-gemba,
    title = "{GEMBA} V2: Ten Judgments Are Better Than One",
    author = "Junczys-Dowmunt, Marcin",
    editor = "Haddow, Barry  and
      Kocmi, Tom  and
      Koehn, Philipp  and
      Monz, Christof",
    booktitle = "Proceedings of the Tenth Conference on Machine Translation",
    month = nov,
    year = "2025",
    address = "Suzhou, China",
    publisher = "Association for Computational Linguistics",
    url = "https://aclanthology.org/2025.wmt-1.67/",
    doi = "10.18653/v1/2025.wmt-1.67",
    pages = "926--933",
    ISBN = "979-8-89176-341-8"
}

@inproceedings{juraska-etal-2024-metricx,
    title = "{M}etric{X}-24: The {G}oogle Submission to the {WMT} 2024 Metrics Shared Task",
    author = "Juraska, Juraj  and
      Deutsch, Daniel  and
      Finkelstein, Mara  and
      Freitag, Markus",
    editor = "Haddow, Barry  and
      Kocmi, Tom  and
      Koehn, Philipp  and
      Monz, Christof",
    booktitle = "Proceedings of the Ninth Conference on Machine Translation",
    month = nov,
    year = "2024",
    address = "Miami, Florida, USA",
    publisher = "Association for Computational Linguistics",
    url = "https://aclanthology.org/2024.wmt-1.35/",
    doi = "10.18653/v1/2024.wmt-1.35",
    pages = "492--504"
}

@inproceedings{rei-etal-2022-cometkiwi,
    title = "{C}omet{K}iwi: {IST}-Unbabel 2022 Submission for the Quality Estimation Shared Task",
    author = "Rei, Ricardo  and
      Treviso, Marcos  and
      Guerreiro, Nuno M.  and
      Zerva, Chrysoula  and
      Farinha, Ana C  and
      Maroti, Christine  and
      C. de Souza, Jos{\'e} G.  and
      Glushkova, Taisiya  and
      Alves, Duarte  and
      Coheur, Luisa  and
      Lavie, Alon  and
      Martins, Andr{\'e} F. T.",
    editor = {Koehn, Philipp  and
      Barrault, Lo{\"i}c  and
      Bojar, Ond{\v{r}}ej  and
      Bougares, Fethi  and
      Chatterjee, Rajen  and
      Costa-juss{\`a}, Marta R.  and
      Federmann, Christian  and
      Fishel, Mark  and
      Fraser, Alexander  and
      Freitag, Markus  and
      Graham, Yvette  and
      Grundkiewicz, Roman  and
      Guzman, Paco  and
      Haddow, Barry  and
      Huck, Matthias  and
      Jimeno Yepes, Antonio  and
      Kocmi, Tom  and
      Martins, Andr{\'e}  and
      Morishita, Makoto  and
      Monz, Christof  and
      Nagata, Masaaki  and
      Nakazawa, Toshiaki  and
      Negri, Matteo  and
      N{\'e}v{\'e}ol, Aur{\'e}lie  and
      Neves, Mariana  and
      Popel, Martin  and
      Turchi, Marco  and
      Zampieri, Marcos},
    booktitle = "Proceedings of the Seventh Conference on Machine Translation (WMT)",
    month = dec,
    year = "2022",
    address = "Abu Dhabi, United Arab Emirates (Hybrid)",
    publisher = "Association for Computational Linguistics",
    url = "https://aclanthology.org/2022.wmt-1.60/",
    doi = "10.18653/v1/2022.wmt-1.60",
    pages = "634--645"
}

@inproceedings{ranasinghe-etal-2020-transquest,
    title = "{T}rans{Q}uest: Translation Quality Estimation with Cross-lingual Transformers",
    author = "Ranasinghe, Tharindu  and
      Orasan, Constantin  and
      Mitkov, Ruslan",
    editor = "Scott, Donia  and
      Bel, Nuria  and
      Zong, Chengqing",
    booktitle = "Proceedings of the 28th International Conference on Computational Linguistics",
    month = dec,
    year = "2020",
    address = "Barcelona, Spain (Online)",
    publisher = "International Committee on Computational Linguistics",
    url = "https://aclanthology.org/2020.coling-main.445/",
    doi = "10.18653/v1/2020.coling-main.445",
    pages = "5070--5081"
}

@inproceedings{lu-etal-2024-error,
    title = "Error Analysis Prompting Enables Human-Like Translation Evaluation in Large Language Models",
    author = "Lu, Qingyu  and
      Qiu, Baopu  and
      Ding, Liang  and
      Zhang, Kanjian  and
      Kocmi, Tom  and
      Tao, Dacheng",
    editor = "Ku, Lun-Wei  and
      Martins, Andre  and
      Srikumar, Vivek",
    booktitle = "Findings of the Association for Computational Linguistics: ACL 2024",
    month = aug,
    year = "2024",
    address = "Bangkok, Thailand",
    publisher = "Association for Computational Linguistics",
    url = "https://aclanthology.org/2024.findings-acl.520/",
    doi = "10.18653/v1/2024.findings-acl.520",
    pages = "8801--8816"
}

@inproceedings{lu-etal-2025-mqm,
    title = "{MQM}-{APE}: Toward High-Quality Error Annotation Predictors with Automatic Post-Editing in {LLM} Translation Evaluators",
    author = "Lu, Qingyu  and
      Ding, Liang  and
      Zhang, Kanjian  and
      Zhang, Jinxia  and
      Tao, Dacheng",
    editor = "Rambow, Owen  and
      Wanner, Leo  and
      Apidianaki, Marianna  and
      Al-Khalifa, Hend  and
      Eugenio, Barbara Di  and
      Schockaert, Steven",
    booktitle = "Proceedings of the 31st International Conference on Computational Linguistics",
    month = jan,
    year = "2025",
    address = "Abu Dhabi, UAE",
    publisher = "Association for Computational Linguistics",
    url = "https://aclanthology.org/2025.coling-main.374/",
    pages = "5570--5587"
}

@inproceedings{kocmi-federmann-2023-large,
    title = "Large Language Models Are State-of-the-Art Evaluators of Translation Quality",
    author = "Kocmi, Tom  and
      Federmann, Christian",
    editor = "Nurminen, Mary  and
      Brenner, Judith  and
      Koponen, Maarit  and
      Latomaa, Sirkku  and
      Mikhailov, Mikhail  and
      Schierl, Frederike  and
      Ranasinghe, Tharindu  and
      Vanmassenhove, Eva  and
      Vidal, Sergi Alvarez  and
      Aranberri, Nora  and
      Nunziatini, Mara  and
      Escart{\'i}n, Carla Parra  and
      Forcada, Mikel  and
      Popovic, Maja  and
      Scarton, Carolina  and
      Moniz, Helena",
    booktitle = "Proceedings of the 24th Annual Conference of the European Association for Machine Translation",
    month = jun,
    year = "2023",
    address = "Tampere, Finland",
    publisher = "European Association for Machine Translation",
    url = "https://aclanthology.org/2023.eamt-1.19/",
    pages = "193--203"
}

@inproceedings{kocmi-federmann-2023-gemba,
    title = "{GEMBA}-{MQM}: Detecting Translation Quality Error Spans with {GPT}-4",
    author = "Kocmi, Tom  and
      Federmann, Christian",
    editor = "Koehn, Philipp  and
      Haddow, Barry  and
      Kocmi, Tom  and
      Monz, Christof",
    booktitle = "Proceedings of the Eighth Conference on Machine Translation",
    month = dec,
    year = "2023",
    address = "Singapore",
    publisher = "Association for Computational Linguistics",
    url = "https://aclanthology.org/2023.wmt-1.64/",
    doi = "10.18653/v1/2023.wmt-1.64",
    pages = "768--775"
}

@inproceedings{kim-2025-rubric,
    title = "{RUBRIC}-{MQM} : Span-Level {LLM}-as-judge in Machine Translation For High-End Models",
    author = "Kim, Ahrii",
    editor = "Rehm, Georg  and
      Li, Yunyao",
    booktitle = "Proceedings of the 63rd Annual Meeting of the Association for Computational Linguistics (Volume 6: Industry Track)",
    month = jul,
    year = "2025",
    address = "Vienna, Austria",
    publisher = "Association for Computational Linguistics",
    url = "https://aclanthology.org/2025.acl-industry.12/",
    doi = "10.18653/v1/2025.acl-industry.12",
    pages = "147--165",
    ISBN = "979-8-89176-288-6"
}

@inproceedings{burchardt-2013-multidimensional,
    title = "Multidimensional quality metrics: a flexible system for assessing translation quality",
    author = "Lommel, Arle Richard  and
      Burchardt, Aljoscha  and
      Uszkoreit, Hans",
    booktitle = "Proceedings of Translating and the Computer 35",
    month = nov # " 28-29",
    year = "2013",
    address = "London, UK",
    publisher = "Aslib",
    url = "https://aclanthology.org/2013.tc-1.6/"
}

@inproceedings{chimoto-bassett-2022-comet,
    title = "{COMET}-{QE} and Active Learning for Low-Resource Machine Translation",
    author = "Chimoto, Everlyn Asiko  and
      Bassett, Bruce A.",
    editor = "Goldberg, Yoav  and
      Kozareva, Zornitsa  and
      Zhang, Yue",
    booktitle = "Findings of the Association for Computational Linguistics: EMNLP 2022",
    month = dec,
    year = "2022",
    address = "Abu Dhabi, United Arab Emirates",
    publisher = "Association for Computational Linguistics",
    url = "https://aclanthology.org/2022.findings-emnlp.348/",
    doi = "10.18653/v1/2022.findings-emnlp.348",
    pages = "4735--4740"
}

@misc{xu2026rubricasexpertscasespecificmqmrubrics,
      title={Rubric-as-Experts: Case-Specific MQM Rubrics for Translation Quality Evaluation}, 
      author={Weilu Xu and Yunzhi Shen and Xinye Wang and Ranfei Dang and Shujian Huang},
      year={2026},
      eprint={2606.21559},
      archivePrefix={arXiv},
      primaryClass={cs.CL},
      url={https://arxiv.org/abs/2606.21559}, 
}

@misc{kim-2025-longer-context,
   title={Do LLMs Truly Benefit from Longer Context in Automatic Post-Editing?},
   url={http://dx.doi.org/10.36227/techrxiv.176107895.57699371/v1},
   DOI={10.36227/techrxiv.176107895.57699371/v1},
   publisher={Institute of Electrical and Electronics Engineers (IEEE)},
   author={Kim, Ahrii},
   year={2025},
   month=Oct }

\newpage

\clearpage
\onecolumn
\appendix
\crefalias{section}{appendix}
\section*{Appendix}
\newtcolorbox{promptbox}[1]{
  breakable, enhanced,
  colback=white, colframe=black!45, boxrule=0.5pt, arc=2pt,
  left=7pt, right=7pt, top=5pt, bottom=5pt,
  title=#1, coltitle=black, colbacktitle=black!8, fonttitle=\bfseries\small,
  fontupper=\ttfamily\footnotesize,
  before upper={\obeylines\raggedright
    \setlength{\parindent}{0pt}\setlength{\parskip}{1.5pt}
    \everypar{\hangindent=1.4em\hangafter=1}},
  before skip=6pt, after skip=10pt
}

\section{Prompts}
\label{app:prompts}
The three passes use the system prompts below, with \texttt{\{src\_lang\}} and
\texttt{\{tgt\_lang\}} filled per language pair. The severity rubric in
Appendix~\ref{app:severity} is appended verbatim to all three, and each error object in the
JSON output carries an additional field \texttt{"severity": 1-4}. The user message is the
target segment for \flu, the source and target segment wrapped in \texttt{<source>} and
\texttt{<target>} tags for \acc, and the full target document for \cst.

\subsection{\flu Pass}
\begin{promptbox}{System prompt: \flu}
You are a strict proofreader of \{tgt\_lang\}. Find ALL spans that must change to make the text correct, natural \{tgt\_lang\}; scan to the end and report every one. There may be several, or none.
Judge it purely as \{tgt\_lang\} writing, by the norms of \{tgt\_lang\} itself and never another language's. Apply \{tgt\_lang\}-specific phenomena a native writer observes --- e.g. paired ¿ ¡ in Spanish, mandatory kana/kanji and no capitalization in Japanese, sentence-final 。in CJK, German noun capitalization, French spacing before ; : ? !, Arabic/Hebrew script and directionality.
Categories:
- spelling: misspelling, wrong or missing diacritic.
- grammar: ungrammatical construction, wrong inflection, tense, or particle.
- agreement: features that must agree do not (gender, number, case, verb).
- word\_order: an order no fluent writer of \{tgt\_lang\} would use.
- punctuation: wrong or missing punctuation by \{tgt\_lang\} convention.
- orthography: wrong case, spacing, script, or number/date/time/measurement formatting by \{tgt\_lang\} convention (decimal comma, thousands separator, date and time format, unit spacing).
- register: register or variety inconsistent within the text.
- term: a word that is not the established \{tgt\_lang\} term. Do not judge whether a term fits a domain --- that needs the source.
Flag a span only if you are certain it is wrong. If your reason would contain "might", "could", "awkward", "more natural", or "better", it is a preference, not an error --- do not flag it. Report each error separately, even when several occur close together. Do not judge meaning, numbers, or dates.
Output --- JSON: \{"errors": [\{"quote": str, "span": str, "reason": str, "category": str\}]\}
- quote: copy the error with enough surrounding text to locate it uniquely in the text --- roughly 3-6 words, verbatim.
- span: the exact minimal substring within quote that is the error.
- reason: one short clause, in English.
- category: one of the eight above.
- If nothing must change, return \{"errors": []\}.
\end{promptbox}

\begin{promptbox}{User message: \flu}
\{tgt\_seg\}
\end{promptbox}

\subsection{\acc Pass}
\begin{promptbox}{System prompt: \acc}
Check whether the \{tgt\_lang\} target conveys the meaning of the \{src\_lang\} source. Find ALL spans where the target departs from the source in meaning; scan to the end. There may be several, or none.
Judge ONLY meaning. Grammar, spelling, punctuation, and style are checked separately --- do not report them here.
Categories:
- mistranslation: the target says something different from the source. Judge the target word by what the source means HERE, not by whether the target reads plausibly on its own. Cognates and literal renderings that "read fine" (e.g. "flanco" (Spanish) for "flank") are the most-missed errors --- check them.
- wrong\_term: a term valid in general but wrong for what the source refers to.
- addition: content in the target absent from the source (target side).
- hallucination: unsupported invented text with no basis in the source (target side). This is always severity 4 (major).
- omission: content in the source absent from the target (source side).
- untranslated: source-language text left untranslated (target side).
- source\_issue: the source itself is defective (typo, garbled, truncated) and cannot be translated correctly (source side). Only when the source is at fault; a fine source with a wrong target is a mistranslation.
Do not flag paraphrase, reordering, or idiom that preserves meaning, or a difference the target language forces (gender, formality, pronouns).
Flag a span only if you are certain the meaning is wrong. If your reason would contain "might", "could", or "possibly", or if the target is a defensible rendering, it is not an error --- do not flag it.
Output --- JSON: \{"errors": [\{"side": "src"|"tgt", "quote": str, "span": str, "reason": str, "category": str\}]\}
- side: "src" for omission or source\_issue; otherwise "tgt". The quote is found on that side's text.
- quote: copy the error with enough surrounding text to locate it uniquely on its side --- roughly 3-6 words, verbatim.
- span: the exact minimal substring within quote that is the error.
- reason: one short clause --- what the source means vs. what the target says.
- category: one of the seven above.
- If accurate, return \{"errors": []\}.
\end{promptbox}

\begin{promptbox}{User message: \acc}
<source>
\{src\_seg\}
</source>
<target>
\{tgt\_seg\}
</target>
\end{promptbox}

\subsection{\cst Pass}
\begin{promptbox}{System prompt: \cst}
You are given a full \{tgt\_lang\} document. Check whether entities and key terms are rendered CONSISTENTLY across the whole document. Judge only internal consistency, not correctness.
Step 1. List the recurring items that must stay consistent: named entities (people, places, organizations), domain terms, and fixed expressions. Include items referred to across sentences, whether by the same words or by later shortened or pronominal references.
Step 2. For each item appearing more than once, check across the document:
- reference consistency: a single referent must not appear under TWO OR MORE different names/terms that a reader would take as different things (e.g. "McConnell" in one place, "Mitchell" in another). One consistent form across the document is correct, not an error.
- agreement across references: when a later shortened or pronominal form refers back (e.g. a name later taken up as "el coche"), its gender/number and any agreeing words match that chosen form correctly. Apply only if \{tgt\_lang\} marks such agreement.
Flag a span only when one of these breaks. Only flag when you are certain both refer to the identical entity.
Do NOT flag:
- a normal shortened or pronominal reference itself --- that is cohesion; flag only when its agreement breaks grammatically.
- an item that appears only once.
- anything about whether a rendering is correct --- that needs the source.
- an item rendered the SAME way every time --- identical renderings are consistency, never a conflict. Only flag when renderings actually DIFFER.
Output --- JSON:
\{"tracked": [\{"referent": str, "renderings": [str, ...]\}],
\hspace*{1em}"errors": [\{"referent": str, "quote": str, "span": str, "reason": str, "category": "inconsistency"\}]\}
- tracked: EVERY recurring entity or term you checked from Step 1, each with ALL its renderings across the document as they literally appear --- even when they are perfectly consistent. This is your audit trail; do not omit items.
- errors: only referents whose "renderings" list in tracked contains two or more genuinely different forms. If a tracked item has a single rendering, it CANNOT be an error.
\hspace*{1em}- referent: the tracked item this conflict belongs to (same string as in tracked).
\hspace*{1em}- quote: copy the conflicting rendering with enough surrounding text to locate it uniquely in the document --- roughly 3-6 words, verbatim.
\hspace*{1em}- span: the exact minimal substring within quote that is the error.
\hspace*{1em}- reason: name the referent and what it conflicts with.
- If nothing recurs, both lists are empty.
\end{promptbox}

\begin{promptbox}{User message: \cst}
\{tgt\_seg\}
\end{promptbox}

\subsection{Shared Severity Instruction}
\label{app:severity}
The following rubric is appended to all three system prompts, and each error object in the
JSON output carries an additional field \texttt{"severity": 1-4}. The cut point that maps
the scale onto Minor and Major is applied in post-processing, not in the prompt
(Section~\ref{sec:system}).

\begin{promptbox}{Appended to all three system prompts}
For each error, rate its severity on this 1-4 scale:
Scale 1: Slight wording change with no impact on clarity or intent.
Scale 2: Alters wording but message and intent remain mostly clear.
Scale 3: Noticeably distorts meaning or intent; the reader is misled or confused.
Scale 4: Substantial distortion; translation is unfaithful and potentially misleading.

\end{promptbox}

\end{document}